\documentclass{article}
\usepackage{PRIMEarxiv}
\usepackage[utf8]{inputenc} 
\usepackage[T1]{fontenc}    
\usepackage{hyperref}       
\usepackage{url}            
\usepackage{booktabs}       
\usepackage{amsfonts}       
\usepackage{nicefrac}       
\usepackage{microtype}      
\usepackage{lipsum}
\usepackage{graphicx}
\usepackage{xcolor}
\usepackage{arydshln, subcaption}
\usepackage{amsmath}

\graphicspath{{media/}}     

\title{EnGraf-Net: Multiple Granularity Branch Network with Fine-Coarse Graft Grained for Classification Task

\thanks{This is a polished version of the author’s accepted manuscript of a paper published in 
\textit{Computer Analysis of Images and Patterns (CAIP 2021)}, Lecture Notes in Computer Science, vol. 13052. 
The published peer-reviewed version is available online at 
\href{https://doi.org/10.1007/978-3-030-89128-2_38}{https://doi.org/10.1007/978-3-030-89128-2\_38}.}}

\author{
  Riccardo La Grassa \\
  INAF--Astronomical Observatory of Padua\\
  Padova, Italy\\
  \texttt{\{riccardo.lagrassa\}@inaf.it} \\
   \And
  Ignazio Gallo\\
  Department of Theoretical and Applied Sciences\\
  University of Insubria\\
  Varese, Italy\\
  \And
  Nicola Landro \\
  Department of Theoretical and Applied Sciences \\
  University of Insubria\\
  Varese, Italy\\
}

\begin{document}
\maketitle

\begin{abstract}
Fine-grained classification models are designed to focus on the relevant details necessary to distinguish highly similar classes, particularly when intra-class variance is high and inter-class variance is low. Most existing models rely on part annotations—such as bounding boxes, part locations, or textual attributes—to enhance classification performance, while others employ sophisticated techniques to automatically extract attention maps. We posit that part-based approaches, including automatic cropping methods, suffer from an incomplete representation of local features, which are fundamental for distinguishing similar objects. While fine-grained classification aims to recognize the leaves of a hierarchical structure, humans recognize objects by also forming semantic associations. In this paper, we leverage semantic associations structured as a hierarchy (taxonomy) as supervised signals within an end-to-end deep neural network model, termed EnGraf-Net. Extensive experiments on three well-known datasets CIFAR-100, CUB-200-2011, and FGVC-Aircraft demonstrate the superiority of EnGraf-Net over many existing fine-grained models, showing competitive performance with the most recent state-of-the-art approaches, without requiring cropping techniques or manual annotations.
\end{abstract}


\section{Introduction}
In Neuroscience, pattern separation is a process defined as the capability to discriminate a set of similar patterns into less-similar sets of outputs patterns.
In previous studies~\cite{neunuebel2014ca3}, authors provide evidence of pattern separation in Dentate Gyrus (DG) neurons and pattern completion (a complementary process of pattern separation) in CA3 neurons. 
The DG and CA3 regions of the hippocampus have long been hypothesized to mediate these processes, and empirical studies~\cite{neunuebel2014ca3,deshmukh2011representation,neunuebel2013conflicts} provide strong support for this functional dissociation.
Other works~\cite{newman2014ca3}, such as “CA3 Sees the Big Picture while Dentate Gyrus Splits Hairs,” further support this idea and provide additional results corroborating this conclusion. 
Similarly, theoretical models~\cite{madar2019pattern} suggest that the DG performs pattern separation on cortical inputs before transmitting differentiated outputs to CA3. Anatomically, DG is ideally positioned for this function, receiving input via the perforant path from the entorhinal cortex and projecting to CA3. These results provide strong support for long-standing hypotheses that assign distinct roles to hippocampal subregions in neural information processing, setting the stage for further research~\cite{madar2019pattern}.
In deep learning, models separate a main signal (e.g., images, sounds, text) into smaller components using convolutional operations, improving discrimination in pattern recognition tasks. Recently, some works have explored enhancing pattern separation by leveraging semantic associations, such as hierarchical structures or manual/automatic annotations extracted for each image, to improve model performance. Many approaches extract specific crops from images to obtain highly discriminative features~\cite{NTSNET,API-Net,hanselmann2020elope,TASN,CDL}, instead of using all available annotations.

In computer vision, fine-grained classification often relies on hierarchical organizational structures composed of different levels of abstraction, which can be represented as a graph. Nodes closer to the root represent abstract concepts, while deeper nodes correspond to finer-grained abstractions. 
Humans also use hierarchical information to recognize unknown objects; hence, category hierarchies provide rich semantic correlations across multiple levels. During learning, this guidance can regularize the semantic space and help algorithms extract more discriminative features. 
For example, some studies~\cite{labrador} designed models incorporating multiple granularity levels, demonstrating the benefits of hierarchical information. Similarly~\cite{la2021learn}, hierarchical annotations from WordNet were used to construct an end-to-end model that jointly performed fine-grained and hierarchical classification using a simple multi-layer perceptron with three levels of abstraction.
Feature fusion in multi-scale models was introduced in prior works~\cite{lin2017feature}, and recent studies~\cite{liu2018path,ghiasi2019fpn} have highlighted advanced architectures for fusing features at different levels within deep models. These methods utilize lateral connections to perform fusion operations extensively. In our approach, we employ semantic associations as hierarchical supervisory signals to improve pattern recognition.
In fine-grained classification, modern deep models often generate attention maps containing highly discriminative features, which enhances classification performance. However, spatial information (e.g., object context and surrounding environment) also provides useful cues for recognition. For instance~\cite{niche_population}, studies on warbler species show that different species partition resources spatially in a community, demonstrating the importance of environmental context. Similarly, in computer vision, spatial information should not be ignored. In our approach, we retain all spatial information rather than relying on cropped regions.
In this paper, inspired by the pattern separation function of DG neurons in the brain, we simulate DG-like separation in a deep learning model using supervised signals derived from hierarchical semantic information in datasets. We explicitly enforce pattern separation to obtain discriminative features capable of recognizing hierarchical object structures and distinguishing highly similar objects. The scientific contributions of this work are as follows:

\begin{enumerate}
\item \textit{We introduce a Multiple Granularity Branch Network with Fine-Coarse graft grained for Fine-Grained classification task.
Our model termed as EnGraf-Net, uses the hierarchical semantic associations from the datasets to enforce pattern separation and improve discrimination capability.}
\item \textit{We conduct experiments on CIFAR-100, CUB200-2011, and FGVC-Aircraft datasets, demonstrating the effectiveness of our approach relative to baselines and competitive state-of-the-art methods. We also investigate the contribution of each component using ResNet variants through ablation studies. The code and experimental results are publicly released.~\cite{engrafnet_link_repo}}.
\end{enumerate}

\subsection{Related work}
NTS-Net~\cite{NTSNET} introduces a self-supervised mechanism to locate informative regions without using the bounding box or part annotations.
Several works~\cite{xie2013hierarchical,wang2015multiple,jaderberg2015spatial,fu2017look} leverage fine-grained human annotations, such as the location of specific image details, to improve model performance. However, human annotations are expensive and contrary to the deep learning principle that representations should be learned automatically.
NTS-Net uses a Navigator module to automatically localize informative regions, a Teacher module to evaluate the probability that these regions belong to the ground-truth class, and a Scrutinizer module to perform fine-grained classification using these regions. The model selects the top-M informative regions, which may be a limitation due to the fixed number of regions considered.
Similarly~\cite{hanselmann2020fine}, other approaches have developed end-to-end localization modules. For example, AffNet and AttNet integrate a localization module that generates attention maps used to predict bounding boxes of discriminative regions. Some methods~\cite{hanselmann2020elope} propose a module that builds attention maps combined with a Global K-Max pooling function to extract a single feature vector describing the image. These models often require multiple separate training runs rather than a fully end-to-end pipeline.
API-Net~\cite{API-Net} introduces an attentive pairwise interaction network for fine-grained classification, inspired by the way humans compare pairs of images to recognize subtle differences. It uses paired images as input and applies a cross-entropy loss with a score-ranking regularization.
In this paper, we compare our approach with recent models achieving strong performance in fine-grained classification. We also conduct extensive ablation studies to analyze the contribution of each component and validate the effectiveness of our method.

\section{Methodology}
\begin{figure}[t]
\begin{center}
   \subfloat[]{\includegraphics[width=.45\linewidth]{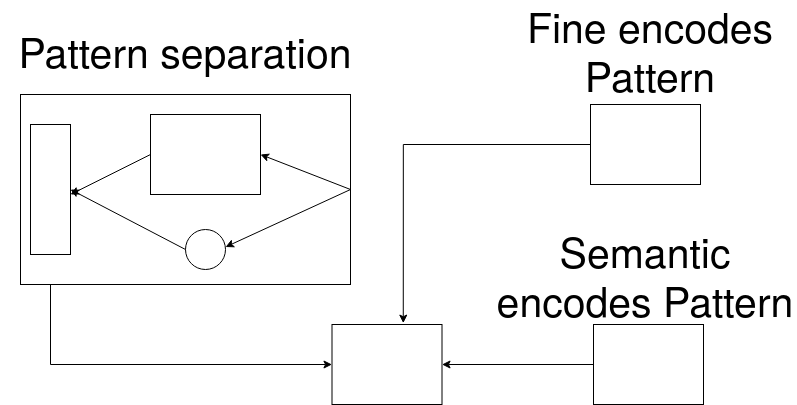}}
   \subfloat[]{\includegraphics[width=.45\linewidth]{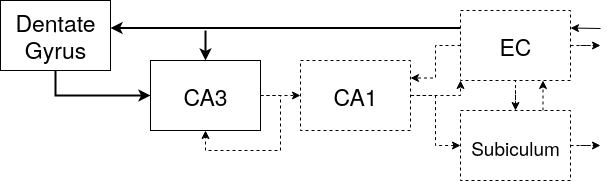}}
\end{center}
   \caption{a) Schematic diagram of our EnGraf-Net b) Schematic diagram of the regions of the hippocampus. The figure shows the feedforward pathway from the entorhinal cortex to the DG and the CA3 neurons. The EC, DG and CA3 blocks are very similar to ours blocks. We simulate the process of the pattern separation by DG and the others connection (EC, DG, CA3) with our proposed approach.}
\label{fig:schematics}
\end{figure}

The hippocampus and its associated structures are capable of transforming similar input patterns into distinct representations (pattern separation) and reconstructing complete stored representations from partial cues (pattern completion).
Figure~\ref{fig:schematics}(b) illustrates the main pathway of the hippocampal regions, along with the structural analogy to our proposed approach (see Fig.~\ref{fig:schematics}(a)). Inspired by this biological process, we simulate pattern separation and completion as a module integrated into a branch of a convolutional neural network and evaluate its performance in fine-grained classification tasks.
Specifically, we enforce a branch through a graft to produce two supervised patterns corresponding to fine labels and coarse labels (pattern separation). These patterns are subsequently concatenated and passed to the next stages of EnGraf-Net, effectively performing pattern completion.
Rather than relying on manual or automatic annotations derived from individual images, we extract semantic associations from the hierarchical structure of the entire dataset. A semantic association quantifies the strength of the relationship between textual units, considering various types of connections, and is widely used in fields such as cognitive psychology and computer science. When organized hierarchically, this structure is referred to as a taxonomy.
We utilize the dataset’s taxonomy, employing semantic associations between classes and superclasses as supervised signals to compute the loss functions. By combining patterns from multiple branches, we enhance the model’s discriminative power and improve overall performance.
Formally, let $y^K$ denote the fine-grained label of a dataset instance. We derive the corresponding superclasses label $y^{K-1}$ from $y^K$.
Each image $x$ is thus annotated with multiple granularity $y^{K-1}, y^K$, where $C_{K-1}, C_{K}$ represent the number of classes at each granularity level.
Our goal is to correctly classify images across these granularity levels using an end-to-end model optimized with cross-entropy loss functions.

\subsection{Network architecture}
EnGraf-Net is built upon the ResNet family of networks.
We adopt a multi-branch architecture (see Fig.~\ref{fig:schematics}), in which the first two branches are designed to extract discriminative features using two types of supervised signals: fine-grained labels for all classes and coarse-grained superclass labels, derived from the dataset’s semantic annotations.
The third branch is dedicated to pattern separation and completion using both supervised signals (fine and coarse grained labels). Different types of grafting can be applied in this branch. Each graft block consists of a convolutional layer, followed by batch normalization and ReLU activation function.
Then, we use an adaptive max pooling with output $1\times 1$, and finally after flattening, a fully connected layer is applied, where $y^{K-1}$ represents the hierarchical class labels.
Outputs from all branches are concatenated and passed through additional fully connected layers, where the final loss function is applied. The total number of parameters in EnGraf-Net depends on the specific ResNet backbone selected, with deeper models resulting in higher parameter counts.

\begin{figure}[t]
\begin{center}
   \includegraphics[width=.5\linewidth]{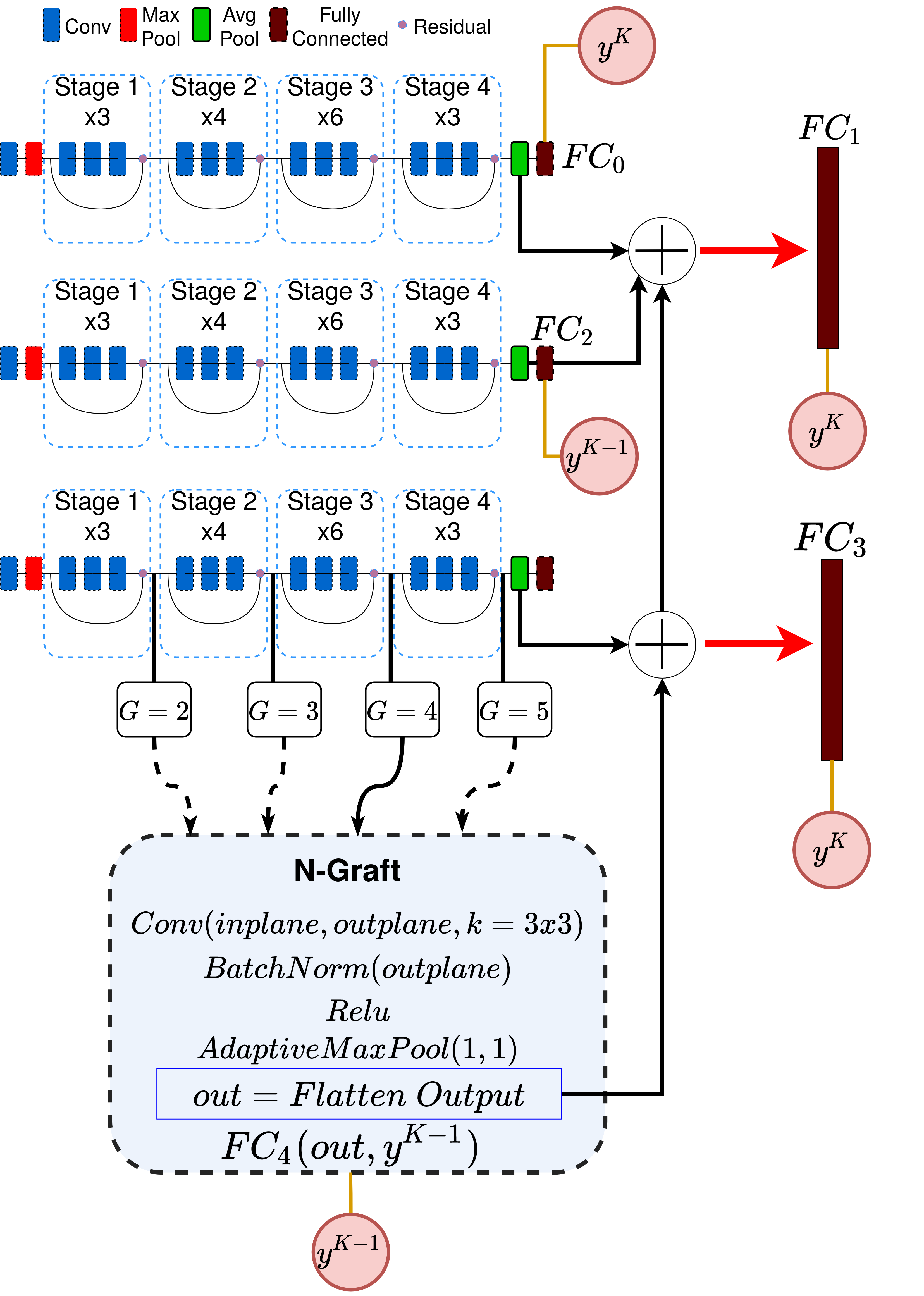}
\end{center}
   \caption{Overview of proposed EnGraf-Net model. It employs two branches to extract features at different grain and a third branch network where we engraft a sub-network useful to apply the pattern separation process.
   }
\label{fig:EnGraf-Net}
\end{figure}

\subsection{Loss functions}
In the training process we use CE as loss function in the form:
\begin{equation}
\mathcal{L}_{xent} = -\frac{1}{m} \sum_{i=1}^{m} \log \frac{e^{W_{y_{i}}^{T}x_i+b_{yi}}}{\sum_{j=1}^{n} e^{W_{j}^{T}x_i+b_j}},
\label{cross-entropy-with-softmax_loss}
\end{equation}
where $W_{yi}$ is the weight associated to class $y$ of i-th instance, $x_i$ are the deep feature of i-th instance and $b$ is the bias term to class $y$ of i-th instance.

Considering the network proposed (see Fig.~\ref{fig:EnGraf-Net}) we compute multiple CE loss in a different part of our proposal ($FC_0, FC_1, FC_2, FC_3, FC_4$) where each of them jointly with supervised signals is used in the learning process with Stochastic Gradient Descendent method to achieve the global minima (or a good approximation of it).
To summarize our total loss function, we use the following formulation:

\begin{align}
\mathcal{L} = & \mathcal{L}_{xent}(FC_0, y^K) + \mathcal{L}_{xent}(FC_1, y^K) + \\ &   \mathcal{L}_{xent}(FC_2, y^{K-1}) +  \mathcal{L}_{xent}(FC_3, y^K) + \mathcal{L}_{xent}(FC_4, y^{K-1}) &  \nonumber
	\label{final_loss}
\end{align}

The cardinality of the classes $y$ considered in \textit{EnGraf-Net} is different in $FC_2, FC_4$ than $FC_0, FC_1, FC_3$ due to the supervised signals selected (it depends on the datasets and by how many hierarchy annotations we consider).

\begin{table}[t]
\caption{Experimental results}
\label{tab:fgvc_abla}
\centering
\resizebox{.9\textwidth}{!}{
\subfloat[CUB-200-2011]{\begin{tabular}{lcc}
\hline\noalign{\smallskip}
Method & Top-1 \\
\hline\noalign{\smallskip}
& \textit{Prior Work}  \\
Resnet-50 &84.5  \\
PN-DCN~\cite{pndcn}(BMVA 14) &85.4 \\
DT-RAM~\cite{dtram}(ICCV 17) &86.0 \\
MC-Loss~\cite{mc_loss}(Trans. Img Proc. 20) &87.3   \\
MaxEnt~\cite{maxEntr}(NeurIPS 18) &86.5  \\
MA-CNN~\cite{zheng2017learning}(ICCV 17) &86.5  \\
KERL~\cite{kerl}(IJCAI 18)    &87.0  \\ 
AP-CNN 1 st.~\cite{ding2021ap}(Trans. Img Proc. 21) &87.2  \\
NTS-Net~\cite{NTSNET}(ICCV 18) &87.5  \\
DBTNet-50~\cite{NEURIPS2019}(NeurIPS 19) &87.5   \\
Cross-X~\cite{luo2019cross}(ICCV 19) &87.7   \\
TASN~\cite{TASN}(CVPR 19)    &87.9  \\ 
HSE~\cite{chen2018fine}(ACM-MM 18) &88.1   \\
DBTNet-101~\cite{NEURIPS2019}(NeurIPS 19) &88.1   \\
CDL~\cite{CDL}(ACM-MM 19)    &88.4  \\
AP-CNN 2 st.~\cite{ding2021ap}(Trans. Img Proc. 21) &88.4  \\
Elope~\cite{hanselmann2020elope}(WACV 20) &88.5  \\
API-Net~\cite{API-Net}(AAAI 20) &88.6  \\
\noalign{\smallskip}\hdashline
& \textit{Our Results} & \\
EnGraf-Net50 (G=4, H=1)  & 87.94  \\
EnGraf-Net101 (G=4, H=1) & 88.00  \\
EnGraf-Net152 (G=4, H=1) & 88.31  \\
\hline
\end{tabular}}
\subfloat[FGVC-Aircraft]{\begin{tabular}{lcc}
\hline\noalign{\smallskip}
Method & Top-1 \\
\hline\noalign{\smallskip}
& \textit{Prior Work} & \\
Kernel-Act~\cite{kernel_acti}(ICCV 17) &88.3  \\
MaxEnt~\cite{maxEntr}(NeurIPS 18) &89.8  \\
MA-CNN~\cite{zheng2017learning}(ICCV 17)  &89.9   \\
PA-CNN~\cite{pacnn}(Trans. Img Proc. 19) &91.0  \\
DBTNet-50~\cite{NEURIPS2019}(NeurIPS 19) &91.2  \\
NTS-Net~\cite{NTSNET}(ICCV 18) &91.4  \\
iSQRT-COV~\cite{li2018towards}(CVPR 18) &91.4  \\
DBTNet-101~\cite{NEURIPS2019}(NeurIPS 19)  &91.6  \\
DFL-CNN~\cite{DFL_cnn}(CVPR 18) &92.0  \\
SEF~\cite{sef}(IEEE Sign. Proc. Lett. 20) &92.1  \\
AP-CNN 1 st~\cite{ding2021ap}(Trans. Img Proc. 21) &92.2  \\
Cross-X~\cite{luo2019cross}(ICCV 19)   &92.7  \\
S3Ns~\cite{s3ns}(ICCV 19) &92.8  \\
MC-Loss~\cite{mc_loss}(Trans. Img Proc. 20) &92.9  \\
EfficientNet-B7~\cite{tan2019efficientnet}(ICML 19) &92.9  \\
API-Net~\cite{API-Net}(AAAI 20) &93.4  \\
Elope~\cite{hanselmann2020elope}(WACV 20)  &93.5   \\
AP-CNN 2 st~\cite{ding2021ap}(Trans. Img Proc. 21)  &94.1  \\
\noalign{\smallskip}\hdashline
& \textit{Our Results}  \\
EnGraf-Net50 (G=4, H=1) &92.14  \\
EnGraf-Net101 (G=4, H=1) & 93.34  \\
\hline
\end{tabular}}}
\resizebox{1.\textwidth}{!}{
\subfloat[Hierarchy classification]{\begin{tabular}{lcc}
\hline\noalign{\smallskip}
& CUB & AIR \\
Method & acc {\color{red}coarse}-{\color{blue}fine} & acc {\color{red}coarse}-{\color{blue}fine} \\
\hline\noalign{\smallskip}
EnGraf-Net50  &{\color{red}92.32}-{\color{blue}87.94}  &{\color{red}95.44}-{\color{blue}92.14} \\
EnGraf-Net101 &{\color{red}92.70}-{\color{blue}88.00}  &{\color{red}96.10}-{\color{blue}93.34}  \\
\end{tabular}
}
\subfloat[Cifar-100]{\begin{tabular}{lcc}
\hline
Method & top-1 \\
\hline\hline
Resnet-18          &72.43 \\
Two-Branch         &72.95 \\
Graft              &73.85 \\
EnGraf-net18 (G=2, H=1) &75.52 \\
EnGraf-net18 (G=3, H=1) &75.13 \\
EnGraf-net18 (G=4, H=1) &\textbf{75.85} \\
EnGraf-net18 (G=5, H=1) &75.41 \\
\hline
\end{tabular}}
\subfloat[Cifar-100]{\begin{tabular}{lccc}
\hline
Method & top-1 & Ours & top-1 \\
\hline\hline
Resnet-18  & 72.43  & EnGraf-net18  & 75.85 \\
Resnet-50  & 75.42  & EnGraf-net50  & 77.27 \\
Resnet-101 & 75.49  & EnGraf-net101 & 77.13 \\
\hline
\end{tabular}}
}
\end{table}

\section{Experiments}
We conducted experiments on three widely used datasets: CIFAR-100, CUB-200-2011, and FGVC-Aircraft. Our goal was to evaluate the performance of EnGraf-Net using different ResNet backbones, comparing our approach against corresponding baselines as well as recent architectures reported in the literature (see Table~\ref{tab:fgvc_abla}a and Table~\ref{tab:fgvc_abla}b).
An ablation study was performed using ResNet-18 to investigate the impact of different graft types and their variations (Table~\ref{tab:fgvc_abla}d and Table~\ref{tab:fgvc_abla}e).
We used Cifar100 dataset as a toy dataset to analyze the behavior of our model. It contains 50,000 images $32\times32$ of training and 10,000 test images, labelled over 100 fine-grained classes.
We use 20 coarse-grained classes as $y^{K-1}$ semantic association in our hierarchical extraction.
All other experiments have been performed on challenging Fine-Grained image classification benchmark datasets.
\textbf{CUB-200-2011}~\cite{WahCUB_200_2011} contains $11788$ images of $200$ species of birds split in $5994$ and  $5794$ images for train and test respectively.
In addition, we use 122 class labelled as \textit{genera} of the species as supervised signals.
\textbf{FGVC-Aircraft}~\cite{maji13fine-grained} contains $10,000$ images of airplanes annotated with the model, specifically splitted in $6667$ and $3333$ for train and test set. This dataset is organised in four-level hierarchy. In addition to 100 classes (fine-labels) we use 70 classes (family) as superclass labelled.
In all our experiments we use different pre-preprocessing data (see our code~\cite{engrafnet_link_repo}).
We report the upper-bound computational time of 19:43h in CUB-200-2011 over $150$ epochs using a learning rate optimizer (SGD in all our experiments) of $0.001$  and batch-size $20$ using an EnGraf-Net152.

\begin{figure}[t]
\centering
   \includegraphics[width=0.25\linewidth]{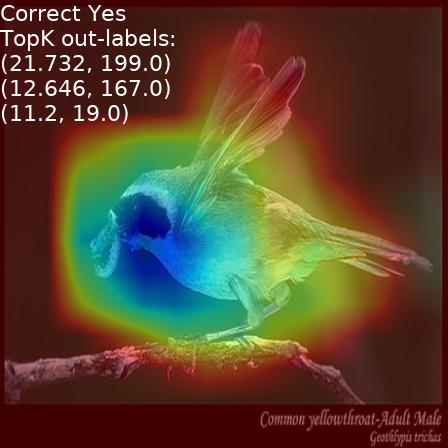}
   \includegraphics[width=0.25\linewidth]{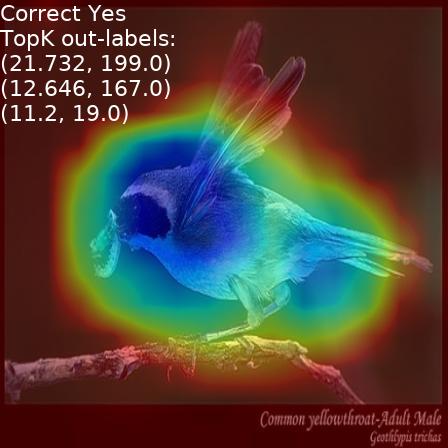}
   \includegraphics[width=0.25\linewidth]{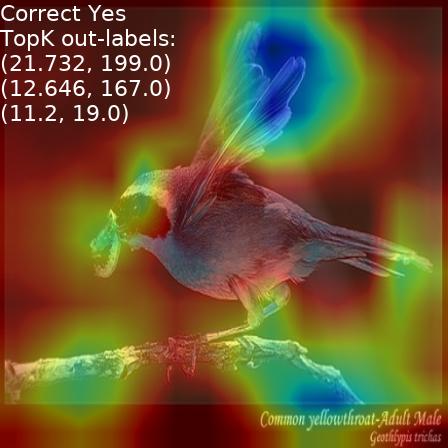}\\
   
   \includegraphics[width=0.25\linewidth]{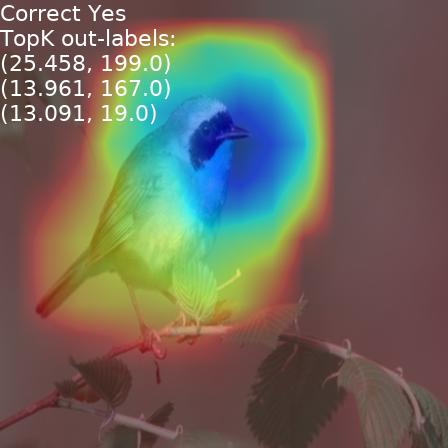}
   \includegraphics[width=0.25\linewidth]{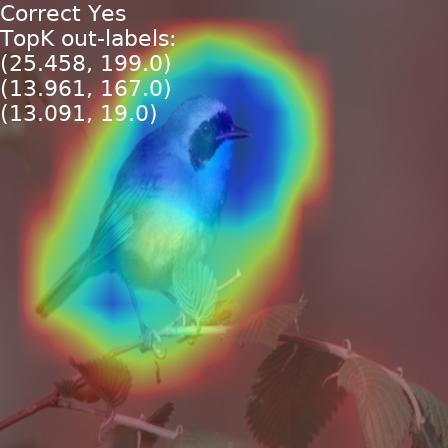}
   \includegraphics[width=0.25\linewidth]{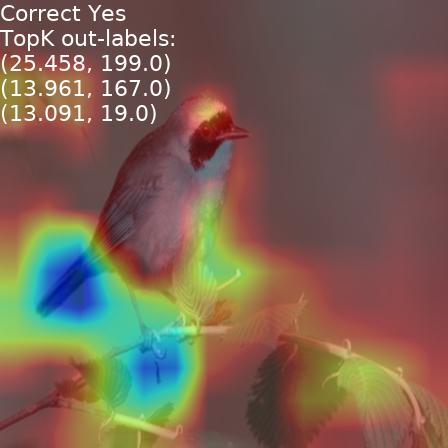}\\
   
   \subcaptionbox{fine-branch}{\includegraphics[width=0.25\linewidth]{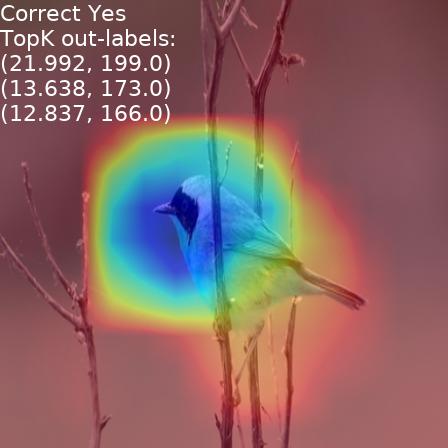}}
   \subcaptionbox{coarse-branch}{\includegraphics[width=0.25\linewidth]{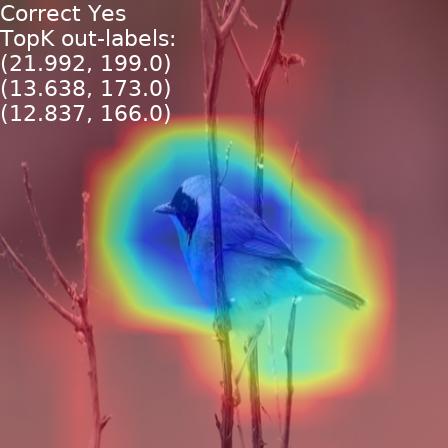}}
   \subcaptionbox{graft-branch}{\includegraphics[width=0.25\linewidth]{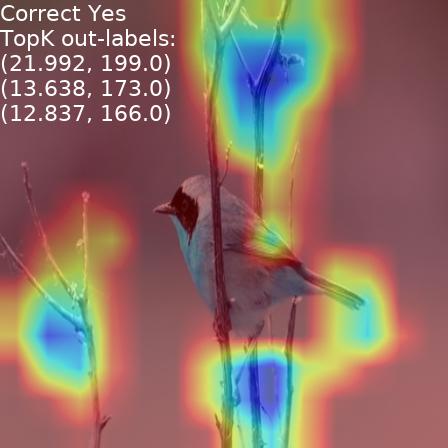}}\\
   
   \caption{Visualization of the attentions regions captured by EnGraf-Net50 in 3 types of layers (columns) and 3 different images (rows) of CUB.
    Using semantic association of the taxonomy, our method has the capability to detect subtle differences and spatial discriminative information without using part annotations.
    The third column show the effectiveness to focus the attention to other regions usually not considered in fine-grained models.}
    \label{fig:feat_visua}
\end{figure}

\subsection{Results}
Tables~\ref{tab:fgvc_abla}a and~\ref{tab:fgvc_abla}b present a comparison between the proposed EnGraf-Net and existing models on two widely used fine-grained classification benchmarks.
Top-1 accuracy was measured in each experiment, demonstrating consistent improvements across both datasets.
Our model achieved $88.31\%$ and $93.34\%$ respectively on CUB-200-2011 and FGVC-Aircraft datasets outperforming the best existing models for fine-grained classification and remaining highly competitive with recent state-of-the-art methods such as API-Net.
We further evaluated performance on the CIFAR-100 dataset to investigate the impact of different graft configurations (Table~\ref{tab:fgvc_abla}d). In these experiments, the best results were obtained with a graft size of 4, outperforming other graft types as well as the baseline ResNet-18, two-branch configurations, and single-branch models with a simple graft. Based on these findings, a graft size of 4 was adopted for all subsequent experiments reported in Tables~\ref{tab:fgvc_abla}a, b, c, and e.
Table~\ref{tab:fgvc_abla}c reports classification accuracy for both coarse and fine-grained classes, demonstrating the ability of EnGraf-Net to solve multiple levels of hierarchical classification in a single end-to-end model.

\subsection{Visualization Analysis}
Several techniques have been proposed in the literature to visualize class activation maps~\cite{selvaraju2017grad,wang2020score}. Among these, Gradient-weighted Class Activation Mapping (Grad-CAM) is widely used because it can be applied to pretrained models and highlights the discriminative regions of images. Such visualization methods are useful for analyzing model behavior and enhancing interpretability.
In Fig.~\ref{fig:feat_visua}, we apply Grad-CAM to visualize the attention maps generated by our model at three different layers. The features extracted from these layers are concatenated and fed into a fully connected layer for final classification.
The first column of Fig.~\ref{fig:feat_visua} shows the activation map from the branch supervised by the fine-grained class label, while the second column shows the branch supervised by the coarse-grained class ($y^{K-1}$). The third column corresponds to the graft branch, which integrates both supervised signals to perform pattern separation and completion.
In conventional fine-grained models, discriminative regions are typically confined to the object itself (as seen in the first column). By contrast, the graft branch identifies additional discriminative regions, including contextual information from the environment and other relevant details (third column). Interestingly, these regions differ noticeably from those highlighted by the other branches.
Our approach effectively explores new spatially informative regions and combines them with features from the other branches, improving overall model performance without requiring any additional annotations such as bounding boxes or part locations.

\section{Conclusions}
In this paper, we proposed a computational model that simulates the pattern separation and completion processes inspired by hippocampal circuits. We introduced a novel approach for fine-grained classification that leverages only semantic associations, without relying on bounding-box or part annotations, nor on sophisticated cropping techniques.
Through extensive experiments using various ResNet backbones, we demonstrated that EnGraf-Net can be integrated into existing convolutional neural networks. Evaluations on CIFAR-100, CUB-200-2011, and FGVC-Aircraft datasets showed that our model consistently outperforms baseline architectures and remains competitive with state-of-the-art methods for fine-grained classification. These results highlight the effectiveness of enforcing hierarchical pattern separation as a strategy to enhance discriminative feature learning in deep neural networks.

\bibliographystyle{unsrt}  
\bibliography{main}

\end{document}